\documentclass[10pt,twocolumn,letterpaper]{article}

\usepackage{iccv}
\usepackage{times}
\usepackage{epsfig}
\usepackage{graphicx}
\usepackage{amsmath}
\usepackage{multirow}
\usepackage{balance}

\usepackage[percent]{overpic}

\usepackage{color}
\definecolor{P}{rgb}{0,0.6,0} 
\definecolor{R}{rgb}{0.5,0,.3} 
\definecolor{J}{rgb}{1,0.5,0} 
\definecolor{V}{rgb}{1,0,0} 

\newcommand{\PM}[1]{ }
\newcommand{\RV}[1]{ }
\newcommand{\JC}[1]{ }
\newcommand{\VM}[1]{ }

\usepackage{amsthm,amsmath,amsfonts,amssymb,textcomp}

\newtheorem{defn}{Definition}

\usepackage[percent]{overpic}

\usepackage[pagebackref=true,breaklinks=true,letterpaper=true,colorlinks,bookmarks=false]{hyperref}

\iccvfinalcopy 

\def\iccvPaperID{1346} 

\ificcvfinal\fi
\begin{document}

\title{Curriculum Dropout}


\author{Pietro Morerio$^{1}$, Jacopo Cavazza$^{1,2}$, Riccardo Volpi$^{1,2}$, Ren\'e Vidal$^{3}$ and Vittorio Murino$^{1,4}$\\ \\
$^{1}$Pattern Analysis \& Computer Vision (PAVIS) -- Istituto Italiano di Tecnologia -- \textit{Genova, 16163, Italy} \\
$^{2}$Electrical, Electronics and Telecommunication Engineering and Naval Architecture Department \\ (DITEN) -- Universit\`{a} degli Studi di Genova --  \textit{Genova, 16145, Italy}\\
$^{3}$Department of Biomedial Engineering -- Johns Hopkins University -- \textit{Baltimore, MD 21218, USA} \\
$^{4}$Computer Science Department -- Universit\`{a} di Verona --  \textit{Verona, 37134, Italy} \\
{\tt\small \{pietro.morerio,jacopo.cavazza,riccardo.volpi,vittorio.murino\}@iit.it, rvidal@cis.jhu.edu}
}

\maketitle



	
	
	
\begin{abstract}	
	Dropout is a very effective way of regularizing neural networks. Stochastically ``dropping out'' units with a certain probability discourages over-specific co-adaptations of feature detectors, preventing overfitting and improving network generalization. Besides, Dropout can be interpreted as an approximate model aggregation technique, where an exponential number of smaller networks are averaged in order to get a more powerful ensemble. In this paper, we show that using a fixed dropout probability during training is a suboptimal choice. We thus propose a time scheduling for the probability of retaining neurons in the network. This induces an adaptive regularization scheme that smoothly increases the difficulty of the optimization problem. This idea of ``starting easy'' and adaptively increasing the difficulty of the learning problem has its roots in curriculum learning and allows one to train better models. Indeed, we prove that our optimization strategy implements a very general curriculum scheme, by gradually adding noise to both the input and intermediate feature representations within the network architecture. Experiments on seven image classification datasets and different network architectures show that our method, named Curriculum Dropout, frequently yields to better generalization and, at worst, performs just as well as the standard Dropout method.
	
\end{abstract}

\section{Introduction}

Since \cite{Alex}, deep neural networks have become ubiquitous in most computer vision applications. The reason is generally ascribed to the powerful hierarchical feature representations directly learnt from data, which usually outperform classical hand-crafted feature descriptors.

As a drawback, deep neural networks are difficult to train because non-convex optimization and intensive computations for learning the network parameters. Relying on availability of both massive data and hardware resources, the aforementioned training challenges can be empirically tackled and deep architectures can be effectively trained in an end-to-end fashion, exploiting parallel GPU computation.

\begin{figure}[t!]
	\includegraphics[width=\columnwidth,keepaspectratio]{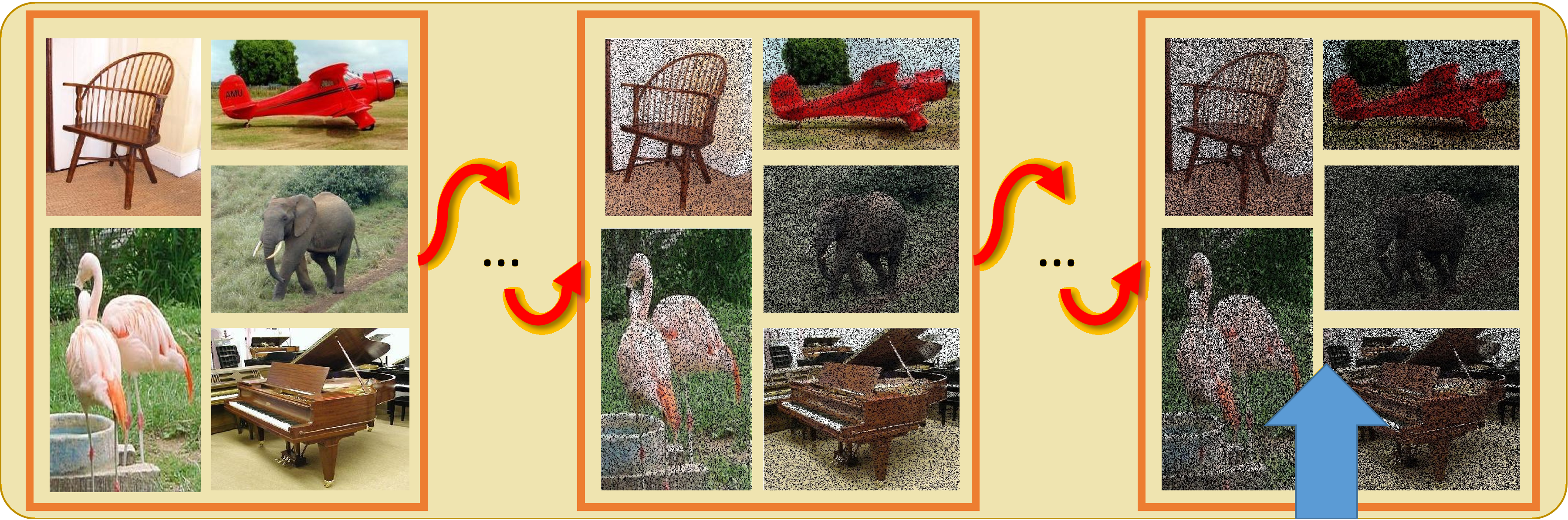}
	\caption{\scriptsize From left to right, during training (red arrows), our curriculum dropout gradually increases the amount of Bernoulli multiplicative noise, generating multiple partitions (orange boxes) within the \textbf{\textit{dataset}} (yellow frame) and the \textbf{\textit{feature representation}} layers (not shown here). Differently, the original dropout \cite{DropoutCORR,DropoutJMLR} (blue arrow) mainly focuses on the hardest partition only, complicating the learning from the beginning and  potentially damaging the network classification performance.}\label{fig:part}
\end{figure}

However, \emph{overfitting} remains an issue. Indeed, such a gigantic number of parameters is likely to produce weights that are so specialized to the training examples that the network's generalization capability may be extremely poor.

The seminal work of \cite{DropoutCORR} argues that overfitting occurs as the result of excessive co-adaptation of feature detectors which manage to perfectly explain the training data. This leads to overcomplicated models which unsatisfactory fit unseen testing data points. To address this issue, the Dropout algorithm was proposed and investigated in \cite{DropoutCORR,DropoutJMLR} and is nowadays extensively used in training neural networks. The method consists in randomly suppressing neurons during training according to the values $r$ sampled from a Bernoulli distribution. More specifically, if $r=1$ that unit is kept unchanged, while if r=0 the unit is suppressed. The effect of suppressing a neuron is that the value of its output is set to zero during the forward pass of training, and its weights are not updated during the backward pass. One one forward-backward pass is completed, a new sample of r is drawn from each neuron, and another forward-backward pass is done and so on till convergence. At testing time, no neuron is suppressed and all activations are modulated by the mean value of the Bernoulli distribution. The resulting model is in fact often interpreted as an average of multiple models, and it is argued that this improves its generalization ability \cite{DropoutCORR,DropoutJMLR}.

Leveraging on the Dropout idea, many works have proposed variations of the original strategy \cite{AdaptiveDropout,AnnealedDropout,DropConv,FastDropout,FastDropoutRecurrent,ImprovedDropout}. However, it is still unclear which variation improves the most with respect to the original dropout formulation \cite{DropoutCORR,DropoutJMLR}. In many works (such as \cite{AnnealedDropout}) there is no real theoretical justification of the proposed approach other than favorable empirical results. Therefore, providing a sound justification still remains an open challenge. In addition, the lack of publicly available implementations (\eg, \cite{ImprovedDropout}) make fair comparisons problematic.

The point of departure of our work is the intuition that the excessive co-adaptation of feature detectors, which leads to overfitting, are very unlikely to occur in the early epochs of training. Thus, Dropout seems unnecessary at the beginning of training. Inspired by these considerations, in this work we propose to dynamically increase the number of units that are suppressed as a function of the number of gradient updates. Specifically, we introduce a generalization of the dropout scheme consisting of a temporal scheduling - a \textit{curriculum} - for the expected number of suppressed units. By adapting in time the parameter of the Bernoulli distribution used for sampling, we smoothly increase the suppression rate as training evolves, thereby improving the generalization of the model.

In summary, the main contributions of this paper are the following.

\begin{enumerate}
	\item We address the problem of overfitting in deep neural networks by proposing a novel regularization strategy called Curriculum Dropout that dynamically increases the expected number of suppressed units in order to improve the generalization ability of the model.  
	
	\item We draw connections between the original dropout framework \cite{DropoutCORR,DropoutJMLR} with regularization theory \cite{citiamolo-pontil-va} and curriculum learning \cite{Bengio:ICML09}. This provides an improved justification of (Curriculum) Dropout training, relating it to existing machine learning methods. 
	
	\item We complement our foundational analysis with a broad experimental validation, where we compare our Curriculum Dropout versus the original one \cite{DropoutCORR,DropoutJMLR} and anti-Curriculum \cite{AnnealedDropout} paradigms, for (convolutional) neural network-based image classification. We evaluate the performance on standard datasets (MNIST \cite{MNIST,Srivastava2015}, SVHN \cite{SVHN}, CIFAR-10/100 \cite{CIFAR}, Caltech-101/256 \cite{Caltech101, Caltech256}). As the results certify, the proposed method generally achieves a superior classification performance. 
\end{enumerate}

The remaining of paper is outlined as follows. Relevant related works are summarized in \S \ref{sez:relwork} and Curriculum Dropout is presented in \S \ref{sez:pip} and \S \ref{sez:CD}, providing foundational interpretations. The experimental evaluation is carried out in \S \ref{sez:exp}. Conclusions and future work are presented in \S \ref{sez:end}.

\section{Related Work}\label{sez:relwork}

As previously mentioned, dropout is introduced by Hinton et al. \cite{DropoutCORR} and Sivrastava et al. \cite{DropoutJMLR}. Therein, the method is detailed and evaluated with different types of deep learning models (Multi-Layer Perceptrons, Convolutional Neural Networks, Restricted Boltzmann Machines) and datasets, confirming the effectiveness of this approach against overfitting. Since then, many works \cite{Wan:ICML13,ImprovedDropout,FastDropout,FastDropoutRecurrent,DropConv,AdaptiveDropout,Wager:NIPS13,AnnealedDropout} have investigated the topic. 

Wan et al. \cite{Wan:ICML13} propose Drop-Connect, a more general version of Dropout. Instead of directly setting units to zero, only some of the network connections are suppressed. This generalization is proven to be better in performance but slower to train with respect to \cite{DropoutCORR,DropoutJMLR}. Li et al. \cite{ImprovedDropout} introduce data-dependent and Evolutional-dropout for shallow and deep learning, respectively. These versions are based on sampling neurons form a multinomial distribution with different probabilities for different units. Results show faster training and sometimes better accuracies. Wang et al. \cite{FastDropout} accelerate dropout. In their method, hidden units are dropped out using approximated sampling from a Gaussian distribution. Results show that \cite{FastDropout} leads to fast convergence without deteriorating the accuracy. Bayer et al. \cite{FastDropoutRecurrent} carry out a fine analysis, showing that dropout can be proficiently applied to Recurrent Neural Networks. Wu and Gu \cite{DropConv} analyze the effect of dropout on the convolutional layers of a CNN: they define a probabilistic weighted pooling, which effectively acts as a regularizer. Zhai and Zhang \cite{Zhai:CoRR15} investigate the idea of dropout once applied to matrix factorization. Ba and Frey \cite{AdaptiveDropout} introduce a binary belief network which is overlaid on a neural network to selectively suppress hidden units. The two networks are jointly trained, making the overall process more computationally expensive. 
Wager et al. \cite{Wager:NIPS13} apply Dropout on generalized linear models and approximately prove the equivalence between data-dependent $L^2$ regularization and dropout training with AdaGrad optimizer. Rennie et al. \cite{AnnealedDropout} propose to adjust the dropout rate, linearly decreasing the unit suppression rate during training, until the network experiences no dropout.

While some of the aforementioned methods can be applied in tandem, there is still a lack of understanding about which one is superior - this is also due to the lack of publicly released code (as happens in \cite{ImprovedDropout}). 
In this respect, \cite{AnnealedDropout} is the most similar to our work. A few papers do not go beyond a bare experimental evaluation of the proposed dropout variation \cite{ImprovedDropout,FastDropoutRecurrent,DropConv,AdaptiveDropout,AnnealedDropout}, omitting to justify the soundness of their approach. Conversely, while some works are much more formal than ours \cite{FastDropout,Wager:NIPS13,Zhai:CoRR15}, all of them rely on approximations to carry out their analysis which is biased towards shallow models (logistic \cite{Wager:NIPS13} or linear regression \cite{FastDropout,Wager:NIPS13} and matrix factorization \cite{Zhai:CoRR15}). Differently, in our paper, in addition to its experimental effectiveness, we provide several natural justifications to corroborate the proposed dropout generalization for deep neural networks.

\section{A Time Scheduling for the Dropout Rate}\label{sez:pip}

Deep Neural Networks display co-adaptations between units in terms of concurrent activations of highly organized clusters of neurons. During training, the latter specialize themselves in detecting certain details of the image to be classified, as shown by Zeiler and Fergus \cite{Zeiler:ECCV14}. They visualize the high sensitivity of certain filters in different layers in detecting dogs, people's faces, wheels and more general ordered geometrical patterns \cite[Fig. 2]{Zeiler:ECCV14}. Moreover, such co-adaptations are highly generalizable across different datasets as proved by Torralba's work \cite{Zhou:NIPS14}. Indeed, the filter responses provided in the AlexNet within \textit{conv1}, \textit{pool2/5} and \textit{fc7} layers are very similar \cite[Fig. 5]{Zhou:NIPS14}, despite the images used for the training are very different: objects from ImageNet versus scenes from Places datasets. 

These arguments support the existence of some \emph{positive} co-adaptations between neurons in the network. Nevertheless, as soon as the training keeps going, some co-adaptations can also be \emph{negative} if excessively specific of the training images exploited for updating the gradients. Consequently, exaggerated co-adaptations between neurons weaken the network generalization capability, ultimately resulting in overfitting. To prevent it, Dropout \cite{DropoutCORR,DropoutJMLR} precisely contrasts those negative co-adaptations. 

The latter can be removed by randomly suppressing neurons of the architecture, restoring an improved situation where the neurons are more ``independent''. This empirically reflects into a better generalization capability \cite{DropoutCORR,DropoutJMLR}.

\paragraph{Network training is a dynamic process.}
Despite the previous interpretation is totally sound, the original Dropout algorithm cannot precisely accommodate for it. Indeed, the suppression of a neuron in a given layer is modeled by a Bernoulli$(\theta)$ random variable\footnote{To avoid confusion in our notation, please note that $\theta$ is the equivalent of $p$ in \cite{DropoutCORR,DropoutJMLR,Wager:NIPS13}, i.e the probability of \textit{retaining} a neuron.}, $0 < \theta \leq 1$. Employing such distribution is very natural, since it statistically models binary activation/inhibition processes. In spite of that, it seems suboptimal that $\theta$ should be \emph{fixed} during the whole training stage. With this operative choice, \cite{DropoutCORR,DropoutJMLR} is actually treating the negative co-adaptations phenomena as uniformly distributed during the whole training time.

\begin{figure}[t!]
	\centering
	\begin{overpic}[width=\columnwidth]{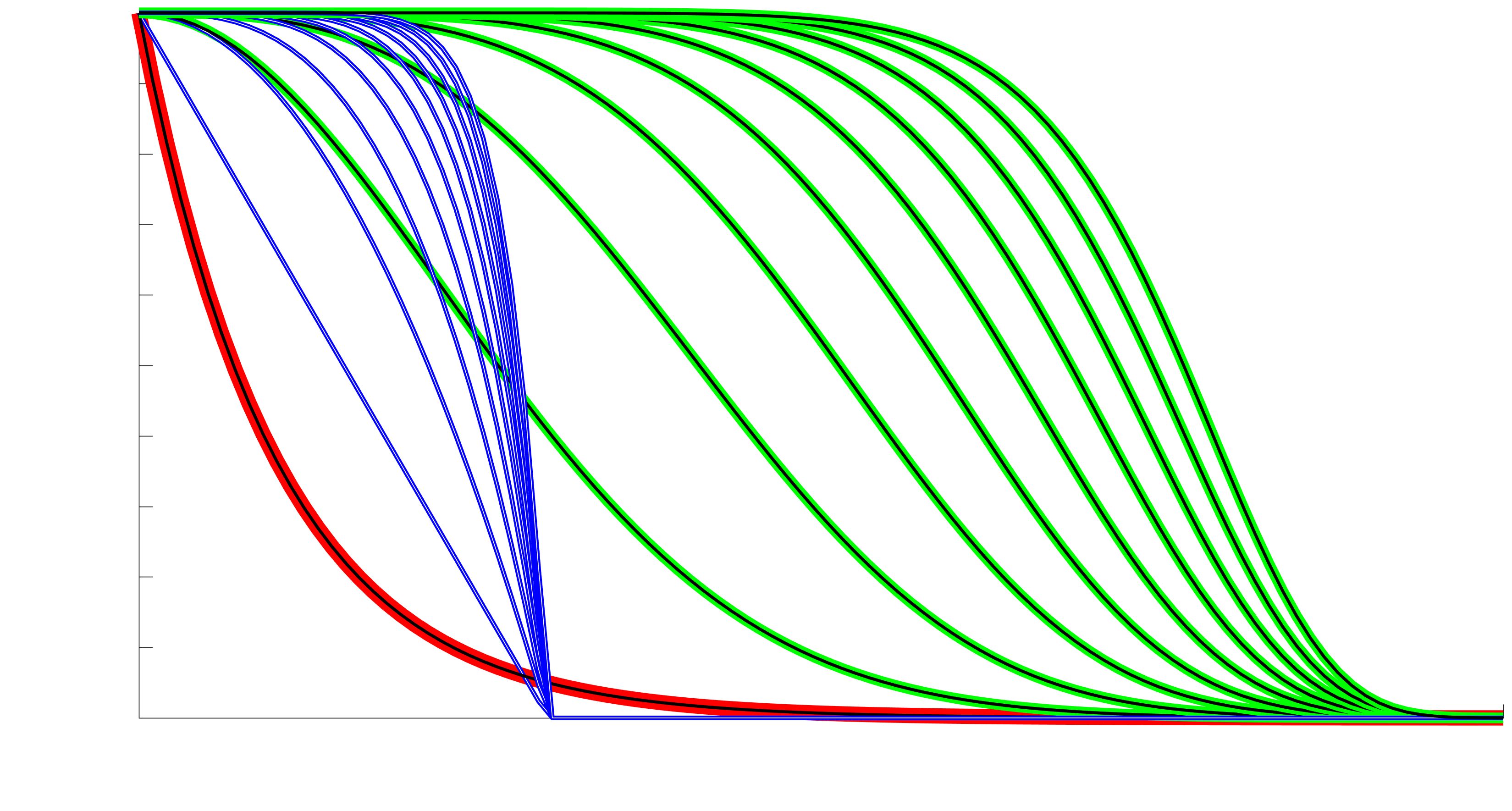}\vspace{1 cm}
		\put (4,0) {$t = 0$}
		\put (85,0) {$t = T$}
		\put (3,5) {$\overline{\theta}$}
		\put (3,51) {$1$}
	\end{overpic}
	\caption{Curriculum functions. Eq. \eqref{eq:sch} (red), polynomial (blue) and exponential (green). }
	\label{fig:curve}
\end{figure}
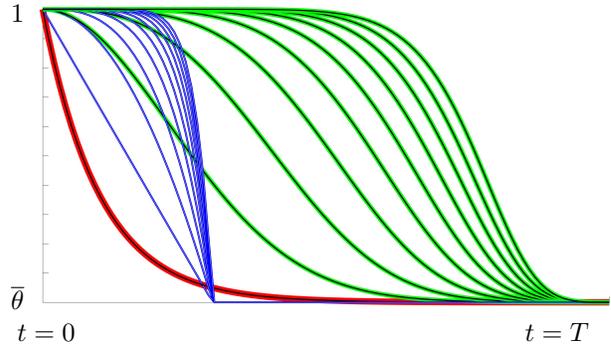 

Differently, our intuition is that, \emph{at the beginning of the training, if any co-adaptation between units is displayed, this should be preserved} as positively representing the self-organization of the network parameters towards their optimal configuration.

We can understand this by considering the random initialization of the network's weights. They are statistically independent and actually not co-adapted at all. Also, it is quite unnatural for a neural network with random weights to overfit the data. On the other hand, the risk of overdone co-adaptations increases as the training proceeds since the loss minimization can achieve a small objective value by overcomplicating the hierarchical representation learnt from data. This implies that \emph{overfitting caused by excessive co-adaptations appears only after a while}.

Since a fixed parameter $\theta$ is not able to handle increasing levels of negative co-adaptations, in this work, we tackle this issue by proposing a temporal dependent $\theta(t)$ parameter. Here, $t$ denotes the training time, measured in gradient updates $t \in \{ 0,1,2,\dots \}$. Since $\theta(t)$ models the probability for a given neuron to be retained, $D \cdot \theta(t)$ will count the average number of units which remain active over the total number $D$ in a given layer. Intuitively, such quantity must be higher for the first gradient updates, then starting decreasing as soon as the training gears. In the late stages of training, such decrease should be stopped. We thus constrain $\theta(t)$ to be $\theta(t) \geq \overline{\theta}$ for any $t$, where $\overline{\theta}$ is a limit value, to be taken as  $0.5 \leq \overline{\theta} \leq 0.9$ as prescribed by the original dropout scheme \cite[\S A.4]{DropoutJMLR} (the higher the layer hierarchy, the lower the retain probability).

Inspired by the previous considerations, we propose the following definition for a \textbf{\textit{curriculum function}} $\theta(t)$ aimed at improving dropout training (as it will become clear in section \ref{sez:CD}, from now on we will often use the terms \textit{curriculum} and \textit{scheduling} interchangeably).

\begin{defn}\label{defn:def}
	Any function $t \mapsto \theta(t)$ such that $\theta(0)=1$ and $\lim_{t \to \infty}\theta(t) \searrow \overline{\theta}$ is said to be a curriculum function to generalize the original dropout \cite{DropoutCORR,DropoutJMLR} formulation with retain probability $\overline{\theta}$.
\end{defn}

Starting from the initial condition $\theta(0)=1$ where no unit suppression is performed, dropout is gradually introduced in a way that $\theta(t) \geq \overline{\theta}$ for any $t$. Eventually (\ie when $t$ is big enough), the convergence $\theta(t) \to \overline{\theta}$ models the fact that we retrieve the original formulation of \cite{DropoutCORR,DropoutJMLR} as a particular case of our curriculum.

Among the functions as in Def. \ref{defn:def}, in our work we fix
\vspace{-.2 cm}
\begin{equation}\label{eq:sch}
	\theta_{\rm curriculum}(t) = (1 - \overline{\theta}) \exp(- \gamma t) + \overline{\theta}, \; \gamma > 0
	\vspace{-.2 cm}
\end{equation} By considering Figure \ref{fig:curve}, we can provide intuitive and straightforward motivations regarding our choice. 

The blue curves in Fig. \ref{fig:curve} are polynomials of increasing degree $\delta = \{1,\dots,10\}$ (left to right). Despite fulfilling the initial constraint $\theta(0) = 1$, they have to be manually thresholded to impose $\theta(t) \to \overline{\theta}$ when $t \to \infty$. This introduces two more (undesired) parameters ($\delta$ and the threshold) with respect to \cite{DropoutCORR,DropoutJMLR}, where the only quantity to be selected is $\overline{\theta}$. 

The very same argument discourages the replacement of the variable $t$ by $t^\alpha$ in \eqref{eq:sch}, (green curves in Fig. \ref{fig:curve}, $\alpha=\{2,\dots,10\}$, left to right). Moreover, by evaluating the area under the curve, we can intuitively measure how aggressively the green curves behave while delaying the dropping out scheme they eventually converge to (as $\theta(t) \to \overline{\theta}$). Precisely, that convergence is faster while moving to the green curves more on the left, being the fastest one achieved by our scheduling function \eqref{eq:sch} (red curve, Fig. \ref{fig:curve}). 

One could still argue that the parameter $\gamma > 0$ is annoying since it requires cross validation. This is not necessary: in fact, $\gamma$ can actually be \emph{fixed} according to the following heuristics. Despite Def. \ref{defn:def} considers the limit of $\theta(t)$ for $t \to \infty$, such condition has to be operatively replaced by $t \approx T$, being $T$ the total number of gradient updates needed for optimization. It is thus totally reasonable to assume that the order of magnitude of $T$ is a priori known and fixed to be some power of $10$ such as $10^4,10^5$. 
Therefore, for a curriculum function as in Def. \ref{defn:def}, we are interested in furthermore imposing $\theta(t) \approx \overline{\theta}$ when $t \approx T$. Actually, a rule of thumb such as
\vspace{-.2 cm}
\begin{equation}\label{eq:heu}
	\gamma = 10/T
	\vspace{-.2 cm}
\end{equation}
implies $|\theta_{\rm curriculum}(T) - \overline{\theta}| < 10^{-4}$ and was used for all the experiments 
in \S \ref{sez:exp}. Additionally, from Figure \ref{fig:curve}, we can grab some intuitions about the fact that the asymptotic convergence to $\overline{\theta}$ is indeed realized for a quite consistent part of the training and well before $t \approx T$. This means that during a big portion of the training, we are actually dropping out neurons as prescribed in \cite{DropoutCORR,DropoutJMLR}, addressing the overfitting issue. In addition to these arguments, we will provide complementary insights on our scheduled implementation for dropout training.
\paragraph{Smarter initialization for the network weights.} The problem of optimizing deep neural networks is non-convex due to the non-linearities (ReLUs) and pooling steps. In spite of that, a few theoretical papers have investigated this issue under a sound mathematical perspective. For instance, under mild assumptions, Haeffele and Vidal \cite{Vidal} derive sufficient conditions to ensure that a local minimum is also a global one to guarantee that the former can be found when starting from \emph{any} initialization. The same theory presented in \cite{Vidal} cannot be straightforwardly applied to the dropout case due to the pure deterministic framework of the theoretical analysis that is carried out. Therefore, it is still an open question whether all initializations are equivalent for the sake of a dropout training and, if not, which ones are preferable. Far from providing any theoretical insight in this flavor, we posit that Curriculum Dropout can be interpreted as a smarter initialization. Indeed, we implement a soft transition between a classical dropout-free training of a network versus the dropout one \cite{DropoutCORR,DropoutJMLR}. Under this perspective, our curriculum seems equivalent to performing dropout training of a network whose weights have already been slightly optimized, evidently resulting in a better initialization for them. 

As a naive approach, one can think to perform regular training for a certain amount of gradient updates and then apply dropout during the remaining ones. We call that \emph{Switch-Curriculum}. This actually induces a discontinuity in the objective value which can damage the performance with respect to the smooth transition performed by our curriculum \eqref{eq:sch} - check Fig. \ref{fig:switch}.

\paragraph{Curriculum Dropout as adaptive regularization.} 

Several connections \cite{Wager:NIPS13,Wan:ICML13,DropoutJMLR,Zhai:CoRR15} have been established between Dropout and model training with noise addition \cite{Bishop:NC95,Rifai:CoRR11,Zhai:CoRR15}. The common trend discovered is that when an unregularized loss function is optimized to fit artificially corrupted data, this is actually \emph{equivalent} to minimize the same loss augmented by a data dependent penalizing term. 
In both \cite[Table 2.]{Wager:NIPS13} for linear/logistic regression and \cite[\S 9.1]{DropoutJMLR} for least squares, it is proved that Dropout induces a regularizer which is scaled 

When $\theta = \overline{\theta}$, the impact of the regularization is just \emph{fixed}, therefore rising potential over- and under-fitting issues \cite{citiamolo-pontil-va}. But, for $\theta = \theta_{\rm curriculum}(t)$, when $t$ is small, the regularizer is set to zero ($\theta_{\rm curriculum}(0) = 1$) and we \emph{do not} perform any regularization at all. Indeed, the latter is simply not necessary: the network weights still have values which are close to their random and statistically independent initialization. Hence, overfitting is unlikely to occur at early training steps. Differently, we should expect it to occur as soon as training proceeds: by using \eqref{eq:sch}, the regularizer is now weighted by
\vspace{-.2 cm}
\begin{equation}
	\theta_{\rm curriculum}(t) ( 1 - \theta_{\rm curriculum}(t)),
	\vspace{-.2 cm}
\end{equation}
which is an increasing function of $t$. Therefore, the more the gradient updates $t$, the heavier the effect of the regularization. This is the reason why overfitting is better tackled by the proposed curriculum. 
Despite the overall idea of an adaptive selection of parameters is not novel for either regularization theory \cite{Hansen94,Crammer:NIPS09,Boyd:2011,robustSSC,Cavazza:Huber} or tuning of network hyper-parameters (e.g. learning rate, \cite{Gulcehre17}), to the best of our knowledge, this is the first time that this concept of  time-adaptive regularization is applied to deep neural networks.

\paragraph{\textit{Compendium}.} Let us conclude with some general comments. We posit that there is no overfitting at the beginning of the network training. Therefore, differently from \cite{DropoutCORR,DropoutJMLR}, we allow for a scheduled retain probability $\theta(t)$ which gradually drops neurons out. Among other plausible curriculum functions as in Def. \ref{defn:def}, the proposed choice \eqref{eq:sch} introduces no additional parameter to be tuned
and implicitly provides a smarter weight initialization for dropout training.

The superiority of \eqref{eq:sch} also relates to $i)$ the smoothly increasingly amount of units suppressed and $ii)$ the soft adaptive regularization performed to contrast overfitting.

Throughout these interpretations, we can retrieve a common idea of smoothly changing difficulty of the training which is applied to the network. This fact can be better understood by finding the connections with Curriculum Learning \cite{Bengio:ICML09}, as we explain in the next section.

\section{Curriculum Learning, Curriculum Dropout}\label{sez:CD}

For the sake of clarity, let us remind the concept of curriculum learning \cite{Bengio:ICML09}. Within a classical machine learning algorithm, all training examples are presented to the model in an unordered manner, frequently applying a random shuffling. Actually, this is very different from what happens for the human training process, that is education. Indeed, the latter is highly structured so that the level of difficulty of the concepts to learn is proportional to the \emph{age} of the people, managing easier knowledge when babies and harder when adults. This ``start small'' paradigm will likely guide the learning process \cite{Bengio:ICML09}.

Following the same intuition, \cite{Bengio:ICML09} proposes to subdivide the training examples based on their difficulty. Then, the learning is configured so that easier examples come first, eventually complicating them and processing the hardest ones at the end of the training. This concept is formalized by introducing a learning time $\lambda \in [0,1]$, so that training begins at $\lambda = 0$ and ends at $\lambda = 1$. At time $\lambda$, $Q_\lambda(z)$ denotes the distribution which a training example $z$ is drawn from. The notion of curriculum learning is formalized requiring that $Q_\lambda$ ensures a sampling of examples $z$ which are easier than the ones sampled from $Q_{\lambda + \varepsilon}$, $\varepsilon > 0$.  Mathematically, this is formalized by assuming
\vspace{-.1 cm}
\begin{equation}\label{eq:www}
	Q_\lambda(z) \propto W_\lambda (z) P(z).
	\vspace{-.1 cm}
\end{equation}
In \eqref{eq:www}, $P(z)$ is the target training distribution, accounting for all examples, both easy and hard ones. The sampling from $P$ is corrected by the factor $0 \leq W_\lambda (z) \leq 1 $ for any $\lambda$ and $z$. The interpretation for $W_\lambda (z)$ is the measure of the difficulty of the training example $z$. The maximal complexity for a training example is fixed to $1$ and reached at the end of the training, \ie $W_1(z) = 1$, \ie $Q_1(z) = P(z)$. The relationship
\vspace{-.1 cm}
\begin{equation}
	W_\lambda (z) \leq W_{\lambda + \varepsilon} (z)
	\vspace{-.1 cm}
\end{equation}
represents the increased complexity of training examples from instant $\lambda$ to $\lambda + \varepsilon$. Moreover, the weights $W_\lambda (z)$ must be chosen in such a way that
\vspace{-.1 cm}
\begin{equation}\label{eq:HH}
	H(Q_\lambda) < H(Q_{\lambda + \varepsilon}),
	\vspace{-.1 cm}
\end{equation}
where Shannon's entropy $H(Q_\lambda)$ models the fact that the quantity of information exploited by the model during training increases with respect to $\lambda$. 

In order to prove that our scheduled dropout fulfills this definition, for simplicity, we will consider it as applied to the input layer only. This is not restrictive since the same considerations apply to any intermediate layer, by considering that each layer trains the feature representation used as input by the subsequent one.

As the images exploited for training, consider the partitions in the dataset including all the (original) clean data and all the possible ways of corrupting them through the Bernoulli multiplicative noise (see Fig. \ref{fig:part}). Let $\pi$ denote the probability of sampling an uncorrupted $d$-dimensional image within an image dataset (nothing more than a uniform distribution over the available training examples). Let us fix the gradient update $t$. The case of sampling a dropped-out $z$ is equivalent to sampling the corresponding uncorrupted image $z_0$ from $\pi$ and then overlapping it with a binary mask $b$ (of size $d$), where each entry of $b$ is zero with probability $1 - \theta(t)$. By mapping $b$ to the number $i$ of its zeros, 
\begin{equation}
	\mathbb{P}[z] = \mathbb{P}[z_0,i] = {d \choose i } (1 - \theta(t))^i \theta(t)^{d - i} \cdot \pi(z_0).
\end{equation}
Indeed, $(1 - \theta(t))^i \theta(t)^{d - i}$ is the probability of sampling \emph{one} binary mask $b$ with $i$ zeros and ${d \choose i }$ accounts for all the possible combinations. Re-parameterizing the training time $t = \lambda T$, we get
\vspace{-.1 cm}
\begin{equation}\label{eq:CLD}
	Q_\lambda(z) = {d \choose i } (1 - \theta(\lambda T))^i \theta(\lambda T)^{d - i} \cdot \pi(z_0).
	\vspace{-.1 cm}
\end{equation} By defining $P(z) = Q_1(z)$ and 
\vspace{-.1 cm}
\begin{equation}\label{eq:Wlambda}
	W_\lambda(z) = \dfrac{1}{P(z)}{d \choose i } (1 - \theta(\lambda T))^i \theta(\lambda T)^{d - i} \cdot \pi(z_0),
	\vspace{-.1 cm}
\end{equation}
one can easily prove 
that the definition in \cite{Bengio:ICML09} is fulfilled by the choice \eqref{eq:CLD} for curriculum learning distribution $Q_\lambda(z)$. 

To conclude, we give an additional interpretation to Curriculum Dropout.
At $\lambda = 0$, $\theta ( 0 ) = 1$ and no entry of $z_0$ is set to zero. This clearly corresponds to the easiest available example, since the learning starts at $t = 0$ by considering all possible available visual information. When $\theta$ start decreasing to $\theta(\lambda T) \approx 0.99$, only 1\% of $z_0$ is suppressed (on average) and still almost all the information of the original dataset $\mathcal{Z}_0$ is available for training the network. But, as $\lambda$ grows, $\theta(\lambda T)$ decreases and a bigger number of entries are set to zero. This complicates the task, requiring an improved effort from the model to capitalize from the reduced uncorrupted information which is available at that stage of the training process.

After all, this connection between Dropout and Curriculum Learning was possible thanks to our generalization through Def. \ref{defn:def}. Consequently, the original Dropout \cite{DropoutCORR,DropoutJMLR} can be interpreted as considering the single specific value $\overline{\lambda}$ such that $\theta(\overline{\lambda} T) = \overline{\theta}$, being $\overline{\theta}$ the constant retain probability on \cite{DropoutCORR,DropoutJMLR}. This means that, as previously found for the adaptive regularization (see \S \ref{sez:pip}), the level of difficulty $W_{\overline{\lambda}}(z)$ of the training examples $z$ is fixed in the original Dropout. This encounters the concrete risk of either oversimplifying or overcomplicating the learning, with detrimental effects on the model's generalization capability. Hence, the proposed method allows to setup a progressive curriculum $Q_\lambda(z)$, complicating the examples $z$ in a smooth and adaptive manner, as opposed to \cite{DropoutCORR,DropoutJMLR}, where such complication is fixed to equal the maximal one from the very beginning (Fig. \ref{fig:part}).

To conclude, let us note that the aforementioned work \cite{AnnealedDropout} proposes a linear \textit{increase} of the retain probability. According to equations (\ref{eq:www}-\ref{eq:HH}) this implements what \cite{Bengio:ICML09} calls an anti-curriculum: this is shown to perform slightly better or worse than the no-curriculum strategy \cite{Bengio:ICML09} and always worse than any curriculum implementation. Our experiments confirm this finding.

\section{Experiments}\label{sez:exp}

In this Section, we applied Curriculum Dropout to neural networks for image classification problems on different datasets, using Convolutional Neural Network (CNN) architectures and Multi-Layer Perceptrons (MLPs)\footnote{Code available at \url{https://github.com/pmorerio/curriculum-dropout}.}. In particular, we used two different CNN architectures: LeNet \cite{LeNet} and a deeper one (conv-maxpool-conv-maxpool-conv-maxpool-fc-fc-softmax), further called CNN-1 and CNN-2, respectively. In the following, we detail the datasets used and the network architectures adopted in each case.

{\bf MNIST} \cite{MNIST} - A dataset of grayscale images of handwritten digits (from 0 to 9), of resolution 28 $\times$ 28. Training and test sets contain 60.000 and 10.000 images, respectively. For this dataset, we used a three-layer MLP, with 2.000 units in each hidden layer, and CNN-1.

{\bf Double MNIST} - This is a static version of \cite{Srivastava2015}, generated by superimposing two random images of two digits (either distinct or equal), in order to generate 64 $\times$ 64 images. The total amount of images are 70.000, with 55 total classes (10 unique digits classes + ${10 \choose 2} = 45$ unsorted couples of digits) . Training and test sets contain 60.000 and 10.000 images, respectively. Training set's images were generated using MNIST training images, and test set's images were generated using MNIST test images. We used CNN-2.

{\bf SVHN} \cite{SVHN} - Real world RGB images of street view house numbering. We used the cropped 32 $\times$ 32 images representing a single digit (from 0 to 9). We exploited a subset of the dataset, consisting in 6.000 images for training and 1.000 images for testing, randomly selected. We used CNN-2 also in this case.

{\bf CIFAR-10} and {\bf CIFAR-100} \cite{CIFAR} - These datasets collect 32 $\times$ 32 tiny RGB natural images, reporting 6000 and 600 elements per each of the 10 or 100 classes, respectively.  In both datasets, training and test sets contain 50.000 and 10.000 images, respectively. We used CNN-1 for both datasets.

{\bf Caltech-101} \cite{Caltech101} - 300 $\times$ 200 resolution RGB images of 101 classes. For each of them, a variable size of instances is available: from 30 to 800. To have a balanced dataset, we used 20 and 10 images per class for training and testing, respectively. Images were reshaped to $128 \times 128$ pixels. We used CNN-2 again here.

{\bf Caltech-256} \cite{Caltech256} - 31000 RGB images for 256 total classes. For each class, we used 50 and 20 images for training and testing, respectively. Images were reshaped to $128 \times 128$ pixels. We used CNN-2.

For training CNN-1, CNN-2 and MLP, we exploited a cross-entropy cost function with Adam optimizer \cite{ADAM} and a momentum term of 0.95, as suggested in \cite{DropoutJMLR}. We used mini-batches of 128 images and fixed the learning rate to be $10^{-4}$. 

\begin{figure*}[ht!]
	\centering
	\begin{overpic}[width=.47\textwidth, keepaspectratio]{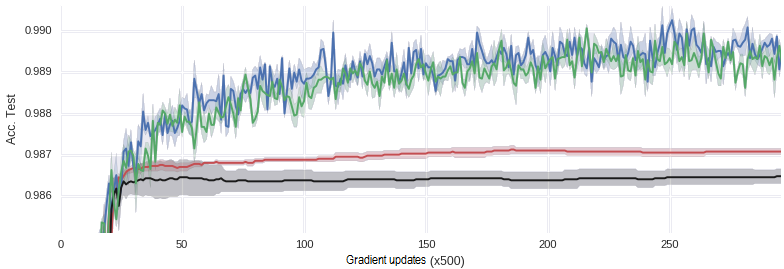}
		\put (10,30) {MNIST \cite{MNIST} (MLP)}
	\end{overpic}
	\begin{overpic}[width=.47\textwidth, keepaspectratio]{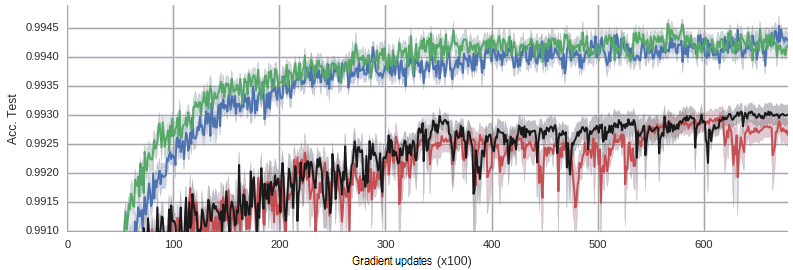}
		\put (8,30) {MNIST \cite{MNIST} (CNN-1)}
	\end{overpic}
	\begin{overpic}[width=.47\textwidth, keepaspectratio]{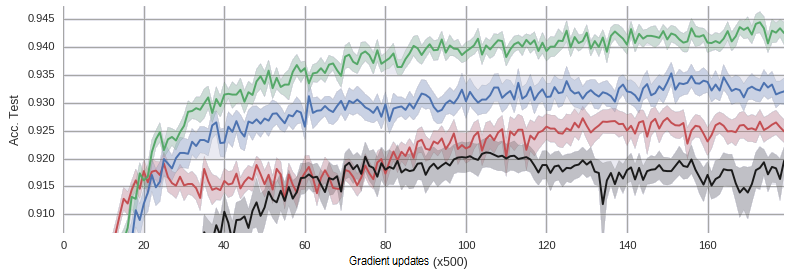}
		\put (10,31) {Double MNIST \cite{Srivastava2015} $ n $ fixed}
	\end{overpic}
	\begin{overpic}[width=.47\textwidth, keepaspectratio]{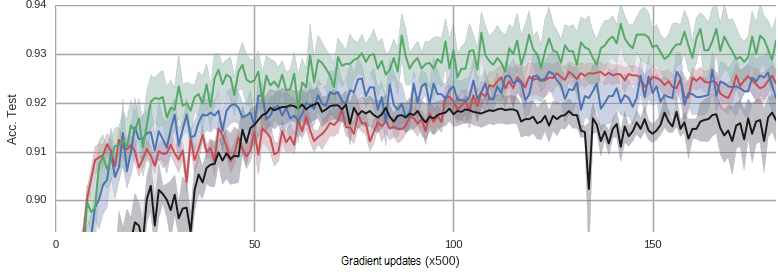}
		\put (8,31) {Double MNIST \cite{Srivastava2015} $ n\overline{\theta} $ fixed}
	\end{overpic}
	\begin{overpic}[width=.47\textwidth, keepaspectratio]{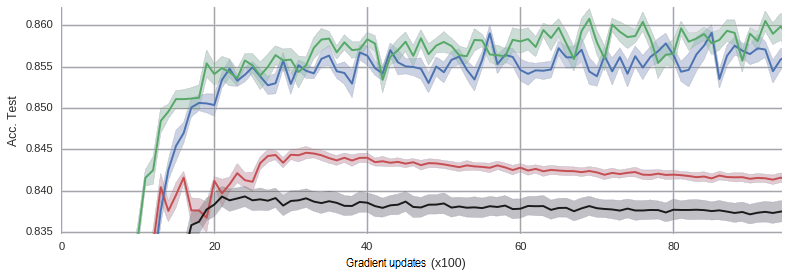}
		\put (10,30) {SVHN \cite{SVHN} $ n $ fixed }
	\end{overpic}
	\begin{overpic}[width=.47\textwidth, keepaspectratio]{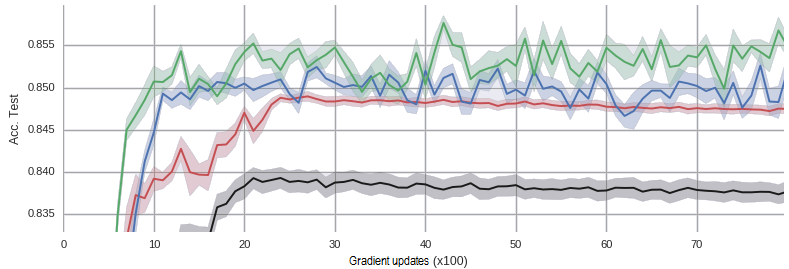}
		\put (8,30) {SVHN \cite{SVHN} $ n\overline{\theta} $ fixed}
	\end{overpic}
	\begin{overpic}[width=.47\textwidth, keepaspectratio]{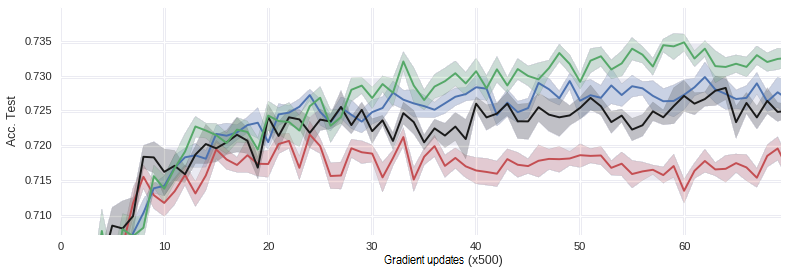}
		\put (10,30) {CIFAR-10 \cite{CIFAR}}
	\end{overpic}
	\begin{overpic}[width=.47\textwidth, keepaspectratio]{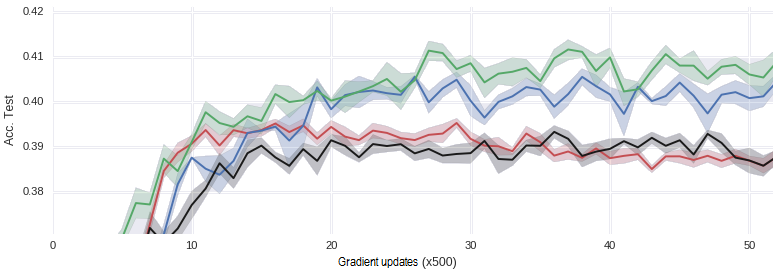}
		\put (8,30) {CIFAR-100 \cite{CIFAR}}
	\end{overpic}
	\begin{overpic}[width=.47\textwidth, keepaspectratio]{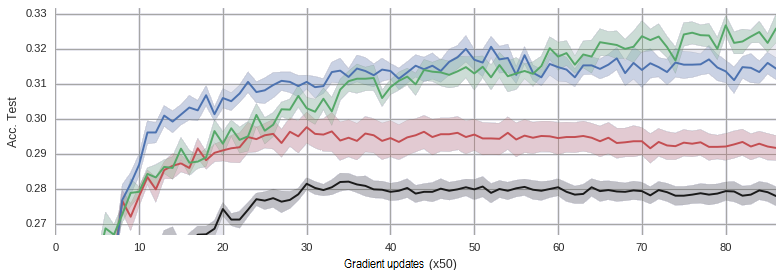}
		\put (10,30) {Caltech-101 \cite{Caltech101}}
	\end{overpic}
	\begin{overpic}[width=.47\textwidth, keepaspectratio]{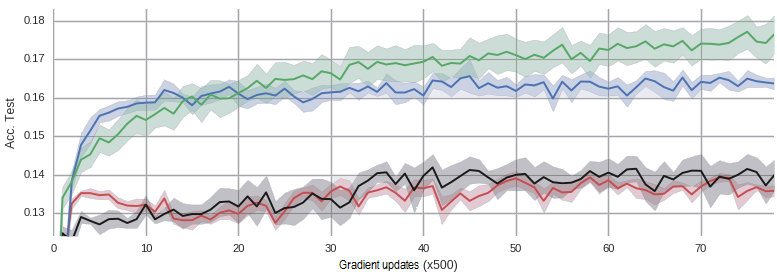}
		\put (8,30) {Caltech-256 \cite{Caltech256}}
	\end{overpic}
	\caption{\scriptsize Curriculum Dropout (green) compared with regular Dropout \cite{DropoutCORR,DropoutJMLR} (blue), anti-Curriculum (red) and a regular training of a network with no units suppression (black). For all cases, we plot mean test accuracy (averaged over 10 different re-trainings) as a function of gradient updates. Shadows represent standard deviation errors.  Best viewed in colors.}
	\label{fig:ResultsAll}
\end{figure*}

\begin{figure*}[h!]
	\centering
	\includegraphics[width = 0.3\linewidth, keepaspectratio]{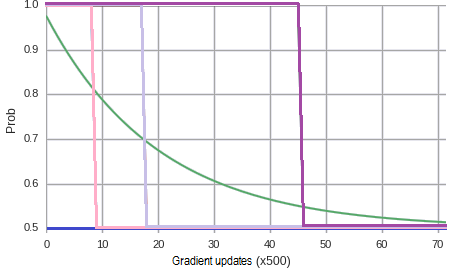}
	\begin{overpic}[width = 0.3\linewidth, keepaspectratio]{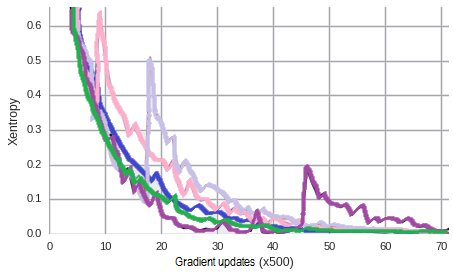}
		\put (42,49) {Double MNIST \cite{Srivastava2015}}
	\end{overpic}
	\begin{overpic}[width = 0.3\linewidth, keepaspectratio]{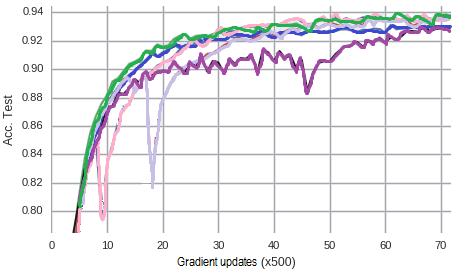}
		\put (42,15) {Double MNIST \cite{Srivastava2015}}
	\end{overpic}
	\caption{\scriptsize Switch-Curriculum. We compare the Curriculum (green) and the regular Dropout (blue) with three cases where we switch from regular to dropout training i) at the beginning (pink) ii) in the middle (violet), iii) almost at the end (purple) of the learning. From left to right, curriculum functions, cross-entropy loss and test accuracy curves. 
	}
	
	\label{fig:switch}
\end{figure*} 

We applied curriculum dropout using the function \eqref{eq:sch} where $\gamma$ is picked using the heuristics \eqref{eq:heu} and $\overline{\theta}$ is fixed as follows.  
For both CNN-1 and CNN-2, the retain probability for the input layer was set to $\overline{\theta}_{\rm input} = 0.9$, selecting $\overline{\theta}_{\rm conv} = 0.75$ and $\overline{\theta}_{\rm fc} = 0.5$ for convolutional and fully connected layers, respectively. For the MLP, $\overline{\theta}_{\rm input} = 0.8$ and $\overline{\theta}_{\rm hidden} = 0.5$. In all cases, we adopted the recommended values \cite[\S A.4]{DropoutJMLR}.  

Before reporting our results, let us emphasize that our 
aim is to improve the standard dropout framework \cite{DropoutCORR,DropoutJMLR}, not to compete for the
state-of-the art performance in image classification tasks. For this reason, we did not use engineering tricks such as data augmentation or any particular
pre-processing, and neither we tried more complex (or deeper) network architectures.

In Fig. \ref{fig:ResultsAll}, we qualitatively compared Curriculum Dropout (green) versus the original Dropout \cite{DropoutCORR,DropoutJMLR} (blue), anti-Curriculum Dropout (red) and an unregularized, \ie no Dropout, training of a network (black). Since CNN-1, CNN-2 and MLP are trained from scratch, in order to ensure a more robust experimental evaluation, we have repeated the weight optimization 10 times for all the cases. Hence, in Fig. \ref{fig:ResultsAll}, we report the mean accuracy value curves, representing with shadows the standard deviation errors. 

Additionally, we report in Table \ref{tab:res} the percentage accuracy improvements of Dropout \cite{DropoutCORR,DropoutJMLR}, anti-Curriculum Dropout \cite{AnnealedDropout} and Curriculum Dropout (proposed) versus a baseline network where no neuron is suppressed. To do that, we selected the average of the 10 highest mean accuracies obtained by each paradigm during each trial; then we averaged them over the 10 runs. We accommodated the metric of \cite{un_bias} to measure the boost in accuracy over \cite{DropoutCORR,DropoutJMLR}. Also, we reproduced for two datasets the cases of fixed layer size $n$ or fixed $n\overline{\theta}$ as in \cite[\S 7.3]{DropoutJMLR}. Here the network layers' size $n$ is preliminary increased by a factor $1/\overline{\theta}$, since on average a fraction $\overline{\theta}$ of the units is dropped out. However, we notice that those bigger architectures tend to overfit the data.

\begin{table}[t!]
	\setlength{\tabcolsep}{1.5pt}
	\footnotesize
	\centering
	\begin{tabular}{|r|c|c|c||c|c|c|cc|}
		\hline
		Dataset & \rotatebox{90}{\scriptsize Architecture} & \rotatebox{90}{\scriptsize Configuration} \rotatebox{90}{\scriptsize ($n$ or $n\overline{\theta} $ fixed)} & \rotatebox{90}{\scriptsize Classes} & \rotatebox{90}{\scriptsize Unregularized}\rotatebox{90}{\scriptsize network}& \rotatebox{90}{\scriptsize  Dropout \cite{DropoutCORR,DropoutJMLR}} & \rotatebox{90}{\scriptsize Anti-Curriculum} & \rotatebox{90}{\scriptsize \textit{Curriculum} }  \rotatebox{90}{\scriptsize \textit{Dropout} }& \rotatebox{90}{\scriptsize (percent boost \cite{un_bias} \vspace{.2 mm} } \rotatebox{90}{\scriptsize $\;$ over Dropout \cite{DropoutCORR,DropoutJMLR}) \hspace{.2 mm}}\\\hline\hline
		\multirow{2}{*}{MNIST \cite{MNIST}} & MLP & $ n $ & \multirow{2}{*}{10} & 98.67 &\textbf{ +0.38} &+0.04 & \textit{+0.36} & \hspace{-4pt}(-5.3\%) \\
		& CNN-1 &$ n $  & & 99.25 & +0.15 &-0.05 & \textbf{\textit{+0.18}} & \hspace{-4pt}\textit{ \textbf{(20.0\%)}} \\ 
		\multirow{2}{*}{{\scriptsize Double} MNIST} & CNN-2 & $ n $ & \multirow{2}{*}{55} & \multirow{2}{*}{92.48} & +1.42 &+0.73 & \textbf{\textit{+2.35}} & \hspace{-4pt}\textit{ \textbf{(65.5\%)}} \\
		& CNN-2 & $ n\overline{\theta} $  &  & & +0.87 &+0.53 & \textbf{\textit{+1.11}} & \hspace{-4pt}\textit{ \textbf{(27.6\%)}} \\
		\multirow{2}{*}{SVHN \cite{SVHN}} & CNN-2 & $ n $  & \multirow{2}{*}{10} & \multirow{2}{*}{84.63} & +2.35 & +1.17& \textbf{\textit{+2.65}} & \hspace{-4pt}\textit{ \textbf{(12.8\%)}} \\ 
		&CNN-2 &$ n\overline{\theta} $  &  &  & +1.59 & +1.51& \textbf{\textit{+2.06}} & \hspace{-4pt}\textit{ \textbf{(29.6\%)}} \\
		CIFAR-10 \cite{CIFAR} & CNN-1 & $ n $  & 10 & 73.06 & +0.22 &-0.68 & \textbf{\textit{+0.62}} & \hspace{-4pt}\textit{ \textbf{(182\%)}} \\
		CIFAR-100 \cite{CIFAR} & CNN-1 & $ n $ & 100 & 39.70 & +1.01 &+0.01  & \textbf{\textit{+1.66}} & \hspace{-4pt}\textit{ \textbf{(64.4\%)}} \\		
		Caltech-101 \cite{Caltech101} & CNN-2 & $ n $  & 101 & 28.56 & +4.21 &+1.57 & \textbf{\textit{+4.72}} & \hspace{-4pt}\textit{ \textbf{(12.1\%)}} \\ 
		Caltech-256 \cite{Caltech256} & CNN-2 & $ n $  & 256 & 14.39 & +2.36 &-0.22 & \textbf{\textit{+3.23}} & \hspace{-4pt}\textit{\textbf{(36.9\%)}} \\ 
		\hline
	\end{tabular}
	
	\vspace{3pt}
	\caption{Comparison of the proposed scheduling versus \cite{DropoutCORR,DropoutJMLR} in terms of percentage accuracy improvement.}\label{tab:res}

\end{table}	

\paragraph{Switch-Curriculum.}
Figure \ref{fig:switch} shows the results obtained on Double MNIST dataset by scheduling the dropout with a step function, \ie no suppression is performed until a certain \textit{switch-epoch} is reached (\S \ref{sez:pip}). Precisely, we switched at 10-20-50 epochs. This curriculum is similar to the one induced by the polynomial functions of Figure \ref{fig:curve}: in fact, both curves have a similar shape and share the drawback of a threshold to be introduced. Yet, Switch-Curriculum shows an additional shortcoming: as highlighted by the spikes of both training and test accuracies, the sudden change in the network connections, induced by the sharp shift in the retain probabilities, makes the network lose some of the concepts learned up to that moment. While early switches are able to recover quickly to good performances, late ones are deleterious. Moreover, we were not able to find any heuristic rule for the \textit{switch-epoch}, which would then be a parameter to be validated. This makes Switch-Curriculum a less powerful option compared to a smoothly-scheduled curriculum.

\paragraph{Discussion.} The proposed Curriculum Dropout, implemented through the scheduling function \eqref{eq:sch}, improves the generalization performance of \cite{DropoutCORR,DropoutJMLR} in almost all cases. As the only exception, in MNIST \cite{MNIST} with MLP, the scheduling is just equivalent to the original dropout framework \cite{DropoutCORR,DropoutJMLR}. Our guess is that the simpler the learning task, the less effective Curriculum Learning. After all, for a task which is relatively easy itself, there is less need for ``starting easy''. This is in any case done at no additional cost nor training time requirements.

As expected, anti-Curriculum was improved by a more significant gap by our scheduling. Also, sometimes, an anti-Curriculum strategy even performs worse than a non-regularized network (\eg, Caltech 256 \cite{Caltech256}). This is coherent with the findings of \cite{Bengio:ICML09} and with our discussion in \S \ref{sez:CD} concerning Annealed Dropout \cite{AnnealedDropout}, of which anti-Curriculum represents a generalization. In addition, while neither regular nor Curriculum Dropout ever need early stopping, anti-Curriculum often does.

\section{Conclusions and Future Work}\label{sez:end}

In this paper we have propose a scheduling for dropout training applied to deep neural networks. By softly increasing the amount of units to be suppressed layerwise, we achieve an adaptive regularization and provide a better smooth initialization for weight optimization. This allows us to implement a mathematically sound curriculum \cite{Bengio:ICML09} and justifies the proposed generalization of \cite{DropoutCORR,DropoutJMLR}.

Through a broad experimental evaluation on 7 image classification tasks, the proposed Curriculum Dropout have proved to be more effective than both the original Dropout \cite{DropoutCORR,DropoutJMLR} and the Annealed \cite{AnnealedDropout}, the latter being an example of anti-Curriculum \cite{Bengio:ICML09} and therefore achieving an inferior performance to our more disciplined approach in ease dropout training. Globally, we always outperform the original Dropout \cite{DropoutCORR,DropoutJMLR} using various architectures, and we improve the idea of \cite{AnnealedDropout} by margin. 

We have tested Curriculum Dropout on image classification tasks only.  However, our guess is that, as standard Dropout, our method is very general and thus applicable to different domains. As a future work, we will apply our scheduling to other computer vision tasks, also extending it for the case of inter-neural connection inhibitions \cite{Wan:ICML13} and Recurrent Neural Networks.

\section*{Acknowledgment}
We gratefully acknowledge the support of NVIDIA Corporation with the donation of one Tesla K40 GPU used for part of this research.

\balance 
{\small
	\bibliographystyle{ieee}
	\bibliography{egbib}
}

\end{document}